\documentclass[11pt]{article}
\usepackage[a4paper, margin=1in]{geometry}
\pdfoutput=1
\usepackage{natbib}
\usepackage{amssymb}
\usepackage{latexsym}
\usepackage{epsfig}
\usepackage{graphicx}
\usepackage[ ruled]{algorithm2e}
\usepackage{algorithmic,  array}
\usepackage{textcomp}
\usepackage{siunitx}
\usepackage{subfig}
\usepackage{wrapfig}
\usepackage[breaklinks=true,bookmarks=false]{hyperref}
\usepackage{url}
\usepackage{xcolor}
\definecolor{newcolor}{rgb}{.8,.349,.1}

\usepackage{mathtools} \usepackage{enumitem,kantlipsum}
\usepackage{cprotect} \usepackage{multirow} \usepackage{collcell}
\usepackage{caption}
\newcommand{\argmax}{\arg\max}
\newcommand{\argmin}{\arg\min}

\usepackage{booktabs}
\DeclareMathAlphabet\mathbfcal{OMS}{cmsy}{b}{n}

\newcolumntype{L}[1]{>{\raggedright\let\newline\\\arraybackslash\hspace{0pt}}m{#1}}
\newcolumntype{C}[1]{>{\centering\let\newline\\\arraybackslash\hspace{0pt}}m{#1}}
\newcolumntype{R}[1]{>{\raggedleft\let\newline\\\arraybackslash\hspace{0pt}}m{#1}}

\title{RC-DARTS: Resource Constrained Differentiable Architecture Search}
\author{Xiaojie Jin$^1$ Jiang Wang$^2$ Joshua Slocum$^2$ Ming-Hsuan Yang$^{2}$ \\ Shengyang Dai$^2$ Shuicheng Yan$^{3}$ Jiashi Feng$^3$\\
\small $^1$Bytedance AI Lab \quad
\small $^2$ Google Inc. \quad
\small$^3$ National University of Singapore}

\begin{document}
\date{}
\maketitle
\begin{abstract}
Recent advances show that Neural Architectural Search (NAS) method
is able to find state-of-the-art image classification deep architectures.
In this paper, we consider the one-shot NAS problem
for resource constrained applications.
This problem is of great interest because it is critical
to choose different architectures according to task complexity when the resource is constrained.
Previous techniques are either too slow for one-shot learning
or does not take the resource constraint into consideration.
In this paper, we propose the resource constrained
differentiable architecture search (RC-DARTS) method to learn architectures that are significantly smaller and faster while achieving comparable accuracy.
Specifically, we propose to formulate the RC-DARTS task as a constrained optimization
problem by adding the resource constraint.
An iterative projection method is
proposed to solve the given constrained optimization problem.
We also propose a multi-level search strategy to enable layers at different depths
to adaptively learn different types of neural architectures.
Through extensive experiments on the Cifar10 and ImageNet datasets, we show that the RC-DARTS method learns lightweight neural architectures which have smaller model size and lower computational complexity while achieving comparable or better performances than the state-of-the-art methods.
\end{abstract}


\section{Introduction}
Deep Neural Networks (DNNs) have demonstrated the state-of-the-art performance on many
machine-learning tasks such as image recognition~\cite{krizhevsky2012imagenet}, speech recognition~\cite{hannun2014deep},
and language modeling~\cite{sutskever2014sequence}.
Despite the successes achieved by DNNs,
crafting neural architectures is usually a time-consuming and costly process that
requires expert knowledge and experience in the field.
Recently, Neural Architecture
Search (NAS) has drawn much attention from both industry and academia~\cite{negrinho2017deeparchitect,zoph2016neural} because
it learns better network automatically from data.
The NAS approaches can be categorized into two main groups.
The methods in the first group
use black-box optimization approaches, such as
Reinforcement Learning (RL)~\cite{zoph2016neural,pham2018efficient,baker2016designing,zoph2017learning,zhong2017practical} or Genetic Algorithm (GA)~\cite{real2017large,xie2017genetic,liu2017hierarchical,real2018regularized},
to optimize a reward function.
The algorithm by Liu et al.  \cite{liu2017progressive} is also a black-box scheme although it uses a slightly more efficient optimization method.
The main drawback of the black-box optimization approaches
is computational cost.
Both RL and GA-based approaches need to train thousands of deep learning models to learn a neural network architecture.
On the other hand, the methods in the second group formulate the neural architecture search task as a differentiable optimization problem and utilizes
alternative gradient descent to find the optimal solution.
One representative example is the  differentiable architecture search (DARTS)~\cite{liu2018darts} method, which has been
shown to perform well on multiple benchmark datasets.
It is also computationally more
efficient than the black-box approaches.

We consider the problem of one-shot NAS which  is critical for resource constrained applications because different tasks require
different neural network architectures.
For example, for a simple problem such as classifying image
color, we can use a simple neural network architecture, e.g.,
a two-layer neural network.
On the other hand, classifying cats and dogs from images
requires a complex neural network.
Pior works on NAS in resource constrained environments is based
on black-box optimization, and  is computationally too expensive for ons-shot NAS.

In this paper, we propose the \textbf{R}esource \textbf{C}onstrained \textbf{N}eural
\textbf{A}rchitecture \textbf{S}earch
(\textbf{RC-DARTS})
for one-shot NAS with good
balance between efficiency and accuracy.
The RC-DARTS method requires the learned architecture to maximize accuracy under user-defined resource constraints.
We uses FLOPs and model size as resource constraints in this paper for simplicity, but can also utilize platform-aware speed as resource constraints by fitting a non-linear mapping function from neural network architecture to inference latency on a device.
Our method is built upon the differentiable architecture search (DARTS)~\cite{liu2018darts} by formulating the problem into a constrained optimization problem by adding resource constraints.
The search space for the resource constrained optimization problem is still a continuous search space, which thus allows for using gradient descent methods.
To solve this optimization problem, we propose an iterative projection algorithm to learn architectures in the feasible set defined by constraints.
We also develop a multi-level search strategy to learn different architectures for layers at different depths, considering that layers at different depths take up distinctive proportions of overall model size and FLOPs.
As such, the proposed RC-DARTS enjoys the merits from the differentiable search space and cost-aware training process.
It is experimentally demonstrated that RC-DARTS learns better lightweight architectures, which are useful in mobile platforms with constrained computing resources.
These properties makes the RC-DARTS approach suitable for one-shot resource constrained NAS problem.

To summarize, we make the following contributions:
\begin{itemize}
	\item We propose an end-to-end resource-constrained NAS framework which is trained in an one-shot manner using standard gradient backpropogation.
	An iterative projection algorithm is introduced to sovle the constrained optimization problem.
	\item We present a multi-level search strategy to learn different architectures for layers at different depths of networks. We also learn a new connection
	cell between adjacent cells.
	It facilitates learning  pareto-optimal architecture across all layers in a network.
	\item We show the proposed RC-DARTS algorithm achieves the state-of-the-arts performance in terms of accuracy, model size and FLOPs on the Cifar10 and ImageNet datasets.
\end{itemize}

\section{Background}
\subsection{Architectural building block}
\label{sec:cell}
Since it is computationally expensive to search for the architecture of the whole network on large-scale dataset,
recent NAS methods~\cite{zoph2017learning,liu2017hierarchical,liu2017progressive,pham2018efficient} usually search for the best architectural building block (or “cell”) on a small-scale dataset.
Multiple building blocks with the same architecture but  independently learned weights are stacked to create a deeper network for larger datasets.

\begin{figure*}
	\centering
	\includegraphics[width=\linewidth]{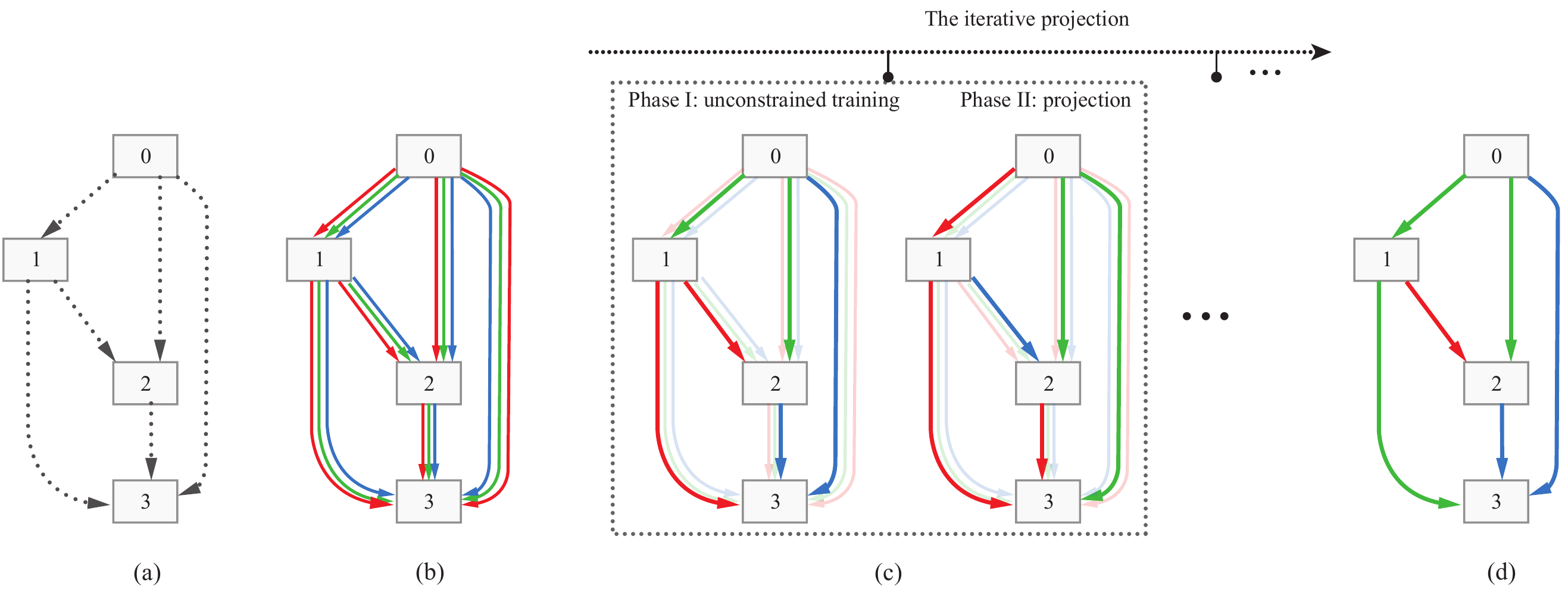}
	\caption{A conceptual visualization of RC-DARTS.
		(a) A cell represented by directed acyclic graph (DAG). The operation corresponding to each edge needs to be learned.
		For clearer illustration, the input/output nodes are omitted.
		(b) Following DARTS~\cite{liu2018darts}, the operation on each edge is replaced by a mixture of all candidate operations (overall 3 candidates in this example).
		The mixing weights for edge $(i,j)$ are parameterized by a learnable vector of $\theta_{( i, j)}$.
		(c) The proposed iterative projection method to solve the constrained optimization problem,
		where the architecture parameters as well as the weights in cell are jointly optimized to satisfy the resource constraints.
		(d) Final architecture derived from the learned weights in mixed operations according to Eq.~\eqref{eq:discrete}.
	}
	\label{fig:overview}
\end{figure*}

As shown in Figure~\ref{fig:overview} (a), a block is represented as a directed acyclic graph (DAG): $G=(V, E)$.
Each node $x_i \in V, i=0, 1, \cdots, N-1$. $x_i$ is an intermediate representation (e.g. a feature map in convolutional networks)
and $N$ is the pre-defined number of nodes in the block.
Corresponding to edge $(i,j)$, there is an operation $O_{i,j}  \in \mathcal{O}$ taking the intermediate representation $x_i$ as inputs and outputting $x_j$.
$\mathcal{O}$ is a pre-defined set of all possible operations including pooling, convolution, zero connection (i.e. no connection between two nodes), etc.
The intermediate representation of a node is the summation of all of its predecessors' transformed outputs:
$
\label{eq:op}
x_j = \sum_{i<j, i\in\mathcal{A}_j}O_{i,j}(x_i),
$ where $\mathcal{A}_j$ is the set of predecessors of $x_j$.
Following ENAS~\cite{zoph2017learning} and DARTS~\cite{liu2018darts},
we define special input nodes and output node for each block.
The first two nodes $x_0$ and $x_1$ in a block are defined as input nodes which transform the outputs of the previous two blocks, respectively.
For the first bock, both $x_0$ and $x_1$ are the input images.
The node $x_{N-1}$ is the output node which is the concatenation
of all intermediate nodes, i.e. $x_{N-1}=\texttt{concat}(x_{2},\cdots, x_{N-2})$.

\subsection{DARTS}
\label{sec:darts}
Since $\mathcal{O}$ is a discrete set, most NAS methods employ time-consuming RL or genetic algorithms to select the best operations in cell.
Recently, DARTS proposes a continuous relaxation to make the architecture search space continuous so that
the architecture can be optimized through gradient descent.
Concretely, the categorical operation $O_{i,j}$ for edge $(i,j)$ is replaced by a mixing operation $\hat{O}_{i,j}$ which outputs the softmax weighted sum of all possible operations in $\mathcal{O}$: $
\hat{O}_{i,j}(x_i) = \sum_{o\in \mathcal{O}} \frac{\exp(\theta_{(i,j)}^o)}{\sum_{o'\in \mathcal{O}}\exp(\theta_{(i,j)}^{o'})}o(x_i)$.
The mixing weights are parameterized by $\theta_{(i,j)} \in \mathbb{R}^{|\mathcal{O}|}$.
The discrete architecture is derived through two steps.
The first step finds $|\mathcal{A}_j|$ strongest predecessors for node $x_j$ based on the \emph{strength} of the corresponding edge.
The strength of an edge $(i,j)$ is defined as $
\max_{o\in\mathcal{O},o\neq \textit{zero}} \frac{\exp(\theta_{(i,j)}^o)}{\sum_{o'\in \mathcal{O}}\exp(\theta_{(i,j)}^{o'})}$.
The second step replaces the mixing operation for
edge $(i,j)$ to a single operation with the largest mixing weight:
\begin{equation}
\label{eq:discrete}
O_{(i,j)}=\argmax_{o\in \mathcal{O},o\neq \textit{zero}}\theta_{(i,j)}^o,\quad \text{s.t.}\quad i \in \mathcal{A}_j.
\end{equation}

The objective of DARTS is to learn neural network architecture by optimizing the following function:
\begin{equation}
\label{eq:darts}
\begin{split}
\min_{\theta} \quad &\mathcal{L}_{\text{val}}(w^*(\theta), \theta)\\
\text{s.t.}  \quad  &w^*(\theta) = \argmin_w \mathcal{L}_{\text{train}}(w, \theta),
\end{split}
\end{equation}
\noindent where $\mathcal{L}_{\text{train}}(w, \theta)$ and $\mathcal{L}_{\text{val}}(w, \theta)$ denote the losses on training and validation dataset respectively, and $w$ are the weights of the operations.
Since it is difficult to get the exact solution for Eq.~\eqref{eq:darts} for both the weights $w$ and the hyperparameters $\theta$ at the same time.
DARTS utilizes coordinate gradient to alternatively update weights $w$ and hyperparameters $\theta$ while fixing the value of the other.

\section{RC-DARTS}
RC-DARTS aims to learn deep architectures under resource constraints such as memory, FLOPs, inference speed, \textit{etc.}
The resource constraints are very important for both mobile platforms and real time applications.
In Section~\ref{sec:rc-obj}, we formulate RC-DARTS as a constrained optimization problem and introduce its objective function.
In Section~\ref{sec:iter_proj}, we introduce the iterative projection method to efficiently optimize the constrained objective function of RC-DARTS.
In Section~\ref{sec:mls}, we propose a multi-level search strategy to adaptively search different architectures for different layers. A new connection cell is introduced to better trade-off between resource costs and accuracy.
The overview of RC-DARTS is illustrated in Figure~\ref{fig:overview}.

\subsection{Objective function of RC-DARTS}\label{sec:rc-obj}
Different from DARTS, RC-DARTS adds resource constraints in objective function,
which require the learned architectures to satisfy task dependent resource constraints, such as the model size and computational complexity.
The objective function of RC-DARTS is
\begin{align}
\min_{\theta} \quad &\mathcal{L}_{\text{val}}(w^*(\theta), \theta) \label{eq:lnas_1} \\
\text{s.t.}  \quad  &w^*(\theta) = \argmin_w \mathcal{L}_{\text{train}}(w, \theta),  \label{eq:lnas_2}\\
& \mathbf{C_L}  \leq \mathbf{\Phi}(\theta)  \leq \mathbf{C_H}, \label{eq:lnas_3}
\end{align}
\noindent where $\mathbf{\Phi}(\theta) = [\Phi^0(\theta), \cdots, \Phi^{M-1}(\theta)]^\top, \Phi^m(\theta):\mathbb{R}^{|\theta|} \mapsto \mathbb{R}$ consists of $M$ cost functions, each of which maps the architecture hyperparameters $\theta$ to a particular resource cost.
$\mathbf{C_L}=[C_L^0, \cdots, C_L^{M-1}]$ and  $\mathbf{C_H}=[C_H^0, \cdots, C_H^{M-1}]$ are
user-defined lower and upper bounds of cost constraints, respectively, i.e.,
the cost $\Phi^m(\theta)$ is constrained to be in the range of $(C_L^m, C_H^m)$.
In this work, we consider two widely used resource costs in Eq.~\eqref{eq:lnas_3} (i.e. $M=2$): the number of parameters and the number of float-point operations (FLOPs).

We introduce the function form of $\Phi^m(\theta)$ as follows.
The exact cost of a neural network architecture can be computed by creating a discretized network architecture
from the network hyper-parameters $\theta$ according to Eq.~\eqref{eq:discrete}, and compute the cost for the discretized network.
However, since the function of discretized network architecture is not continuous,
it is challenging to optimize the objective function with gradient descent.
Similar to DARTS, we use a continuous relaxation strategy on the resource constraints, where the cost of edge $(i,j)$ is calculated as the softmax over all possible operations' costs:

\begin{align}
\label{eq:cost_func}
\Phi^m(\theta) = \sum_{i < j} {\sigma(i\in \mathcal{A}_j)\cdot\mathbf{ u}_{(i,j)}^m}^\top \mathrm{\mathbf{F}}(\theta_{(i,j)})
\end{align}

\noindent where $\mathbf{u_{(i,j)}}$ consists of the resource costs of all operations in $\mathcal{O}$ and $\mathbf{ F}(\cdot)$ is the softmax function. $\sigma(\cdot)$ is the indicator function and $\mathcal{A}_j$ is the set of predecessor nodes for node $j$ (ref. Section~\ref{sec:darts} for the selection of predecessor nodes).
Eq.~\eqref{eq:cost_func} uses the expectation of resource costs in a cell as an approximation to the actual cost of the discrete architecture derived from $\theta$.
There are two advantages to use the function form in Eq.~\ref{eq:cost_func}.
First since Eq.~\ref{eq:cost_func} is differentiable w.r.t. $\theta$, it enables the use of gradient descent to optimize the objective function of RC-DARTS.
Second, it is easy to implement because the resource cost of each candidate operation for the edge $(i,j)$ is independent of the values of $\theta_{(i,j)}$.
Therefore $\mathbf{u}$ is fixed and they can be computed before training.
If people want to optimize a more complicated resource constraint, such as inference speed on a particular platform, we can also
learn a neural network to map from network architecture hyper-parameters to resource cost. We will leave it as our future work.

Note we set both lower bound and higher bound constraints for $\mathbf{\Phi}(\theta)$ to prevent
the model from learning over-simplified architectures.
The lower bound constraints ensures that the model has
sufficient representation capabilities.
Since the cost function w.r.t $\theta$ is non-convex because of the softmax function in Eq.~\eqref{eq:cost_func}, there is no closed-form solution to the objective function.
In the next, we introduce an iterative projection method to optimize the constrained function.

\begin{algorithm}
	\SetKwInOut{Input}{input}\SetKwInOut{Output}{output}
	\caption{Training of RC-DARTS}
	\begin{flushleft}
		Randomly initialized architecture parameters $\theta$ and randomly initialized weights in mixed operations $\hat{O}$.
	\end{flushleft}
	\While{not converged} {
		\emph{Phase I: unconstrained optimization}\\
		\For{$t =1$ to $e_u$} {
			(\textbf{Step 1}) Keep $\theta^t$ fixed, obtain $w^{t+1}$ by descending along $\nabla_{w}\mathcal{L}_{\text{train}}(w^{t}, \theta^t)$\\
			(\textbf{Step 2}) Keep $w^{t+1}$ fixed, obtain $\theta^{t+1}$ by descending along $\nabla_\theta\mathcal{L}_{\text{val}}(w^{t+1}, \theta^t)$.\\
		}
		\emph{Phase II: architectural projection}\\
		(\textbf{Step 3}) Get the projection of $\theta$ outputting at the end of Phase I by solving Eq.~\eqref{eq:project}.\\
		(\textbf{Step 4}) Update architecture parameters with $\theta_p$.
	}
	\label{alg:lnas}
	Following Eq.~\eqref{eq:discrete}, derive the discrete architectures by replacing each mixed operation with the operation with largest mixing weight.
\end{algorithm}

\subsection{Iterative projection}\label{sec:iter_proj}
The proposed iterative projection method optimizes the objective function of RC-DARTS in two alternative phases.
Phase 1 is unconstrained training, which searches better architectures by learning $\theta$ in a larger parameter space without constraints.
Phase II is architecture projection, which projects the network architecture parameters $\theta$ outputting by phase I
to its nearest point in the feasible set defined by constrains in Eq.~\ref{eq:lnas_3}.
The algorithm is shown in Algorithm~\ref{alg:lnas}.

\subsubsection{Unconstrained training}
In this phase, the objective function of RC-DARTS is the same as that of DARTS (i.e. Eq.~\eqref{eq:darts})
because we do not consider the constraints.
As illustrated in Algorithm~\ref{alg:lnas},  $w$ and $\theta$ are alternatively updated for $e_u$ iterations via gradient descend.
Note in step 2, we adopt the simple heuristic of optimizing $\mathcal{L}_{\text{val}}$ by assuming $w$ and $\theta$ are independent, which corresponds to the first-order approximation in DARTS, because it is more computational efficient
while achieving good performance.

There are two benefits of performing phase I.
First, by jointly optimizing weights and architecture, the network learns a good starting point for projection in phase II.
Secondly, after a phase II (note phase I and phase II are performed in an iterative way), phase I is performed to learn architecture in a larger parameter space which has no constraints on $\theta$. Thus, even if a projection step results in a sub-optimal network architecture, the unconstrained
training step can still learn better network architecture based on it.
Besides, it is observed that the network in initial training stage is fragile to perturbations on weights~\cite{yosinski2014transferable}.
In order to improve the training stability, we use a warm-start strategy, i.e. the first phase I in Algorithm~\ref{alg:lnas} has larger $e_u$ than the rest phase I. The warm start reduces the model's risk of getting into bad local optimum in the projection step.

\subsubsection{Architecture projection}
In this phase, we project the architecture hyperparameters $\theta$  of phase I to its nearest point $\theta_p$ in the feasible set defined by Eq.~\eqref{eq:lnas_3}. The objective of projection is
\begin{equation}
\label{eq:project}
\begin{split}
\min_{\theta_p} \quad &\frac{1}{2}||\theta - \theta_p||_2^2\\
\text{s.t.}  \quad &\mathbf{C_L}  \leq \mathbf{\Phi}(\theta_p)  \leq \mathbf{C_H}.
\end{split}
\end{equation}

Because $\mathbf{\Phi}(\theta_p)$ are non-convex functions of $\theta_p$ (ref. to Eq.~\eqref{eq:cost_func}), there is no closed-form solution to Eq.~\eqref{eq:project}.
We transform Eq.~\eqref{eq:project} to its Lagrangian Eq.~\eqref{eq:project_lag} for optimization.

\begin{align}
\label{eq:project_lag}
\small
\min_{\theta_p} \ h(\theta_p) =& \frac{1}{2}||\theta - \theta_p||_2^2 + \lambda_1 \sum_{m=0}^{M}\max(C_L^m - \Phi^m(\theta_p), 0) \nonumber \\
& + \lambda_2\sum_{m=0}^{M} \max(\Phi^m(\theta_p) - C^m_H, 0).
\end{align}

We use regular gradient descent  to optimize Eq.~\eqref{eq:project_lag}.
At time step 0, $\theta_p^0 = \theta$.
At time step $k$, $\theta_p^k$ is obtained by descending $\theta_p^{(k-1)}$ in the direction of $\nabla_{\theta_p}h(\theta_p^{(k-1)})$.
The update is carried out iteratively until all constraints are satisfied or the maximum iteration number $e_p$ is reached.
The output network hyperparameters $\theta_p$ of Eq.~\eqref{eq:project_lag} is utilized as initialization of the next phase I
(i.e. step 4 in Algorithm~\ref{alg:lnas}).

In our experiments, we set the weighting terms of $\lambda_1$ and $\lambda_2$ to be identical for all constraints for simplicity.
To facilitate convergence, we set $\lambda$ to diminish exponentially during training.
At the end of training, $\lambda \rightarrow 0$ and  $\theta_p:=\theta$.
Since it is fast to compute $\nabla_{\theta_p}h(\theta_p)$ for simple resource constraints,
the proposed iterative projection step is much faster than the unconstrained training step.
Thus, RC-DARTS is almost as efficient as regular DARTS.

\subsection{Multi-level architecture search}
\label{sec:mls}
In previous methods~\cite{zoph2017learning,liu2017hierarchical,liu2017progressive,pham2018efficient}, $\theta=(\theta_{\text{normal}}, \theta_{\text{reduce}})$ where $\theta_{\text{normal}}$ is shared by all normal cells, and $\theta_{\text{normal}}$ is shared by all reduce cells.
We argue such a simple solution is sub-optimal for learning resource constrained networks for two reasons.
First, cells at different network depths exhibit large variation on the resource cost (\#params and \#flops),
because the \#channels of filters is increased wherever the resolution of input is reduced.
This design is widely used in current deep networks~\cite{zoph2017learning,liu2017hierarchical,liu2017progressive,pham2018efficient} to avoid the bottleneck of information flow~\cite{szegedy2016rethinking},
where the low-level layers (i.e. layers near input) have larger FLOPs than the high-level layers while the high-level layers have larger number of parameters than low-level layers.
In order to make the learned architectures satisfy given resource constraints, it is better to make the architecture of cells vary with the depths of layers.

Second, cells at different depths have different effects on network's overall performance
For example, it is observed that low-level layers (near input) are more insensitive to reducing the number of parameters~\cite{han2015deep}.

Because of above two reasons, we propose the multi-level search strategy to achieve better trade-off between the architecture's resource costs and accuracy.
Specifically, we evenly divide all layers learned by normal cells into $k$ groups where cells in each group share the same architecture.
In this way, RC-DARTS is encouraged to learn architectures adaptively for different cells.
In addition, to obtain more lightweight architectures, we learn a new type of cell named \textit{connection cell} to learn the candidate connections between cells instead of predefining the connections to be 1x1 conv as in~\cite{zoph2017learning,liu2017hierarchical,liu2017progressive,pham2018efficient}.
The connection cell can be formulated in the same way as normal/reduction cells (ref. Section~\ref{sec:cell}).
The differences are there is only one input node and one intermediate node inside the connection cell. More details on connection cell are introduced in the experiment section.

\section{Experiments}
Similar as DARTS, our experiments consist of three steps. First, RC-DARTS is applied
for searching convolutional cells on CIFAR-10 and the best
cells based on their search validation performance are selected as building blocks in following steps. Then, a larger network is constructed by stacking the
selected cells in previous step and is retrained on CIFAR-10 for comparison between RC-DARTS and other state-of-the-art methods. Finally, we show that the searched convolutional cells are transferable
to large datasets through experimental verification on ImageNet.

\begin{figure*}[t]
	\centering
	\subfloat[][Nomal cell in DARTS\label{fig:darts_normal}]{%
		\includegraphics[width=0.33\linewidth]{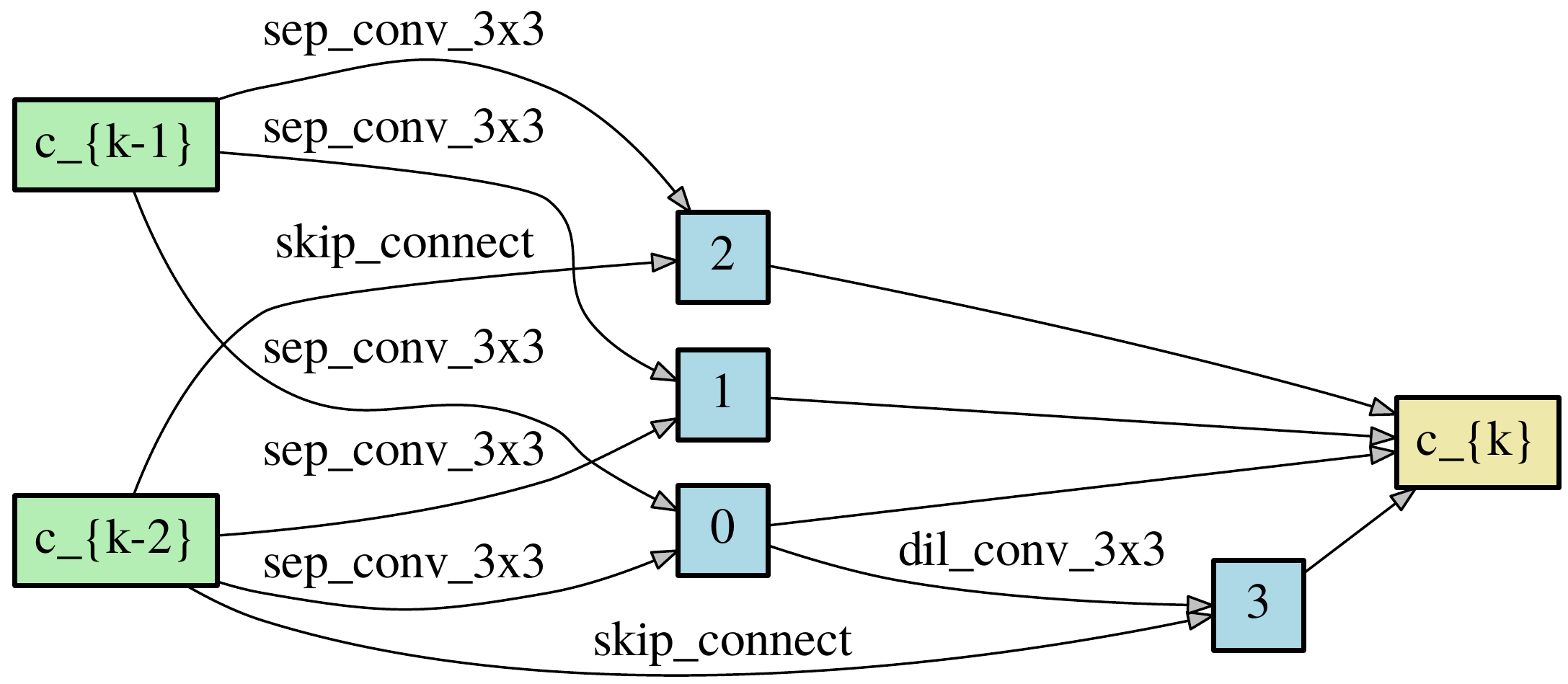}%
	}
	\subfloat[][Reduction cell in DARTS\label{fig:darts_reduce}]{%
		\includegraphics[width=0.33\linewidth]{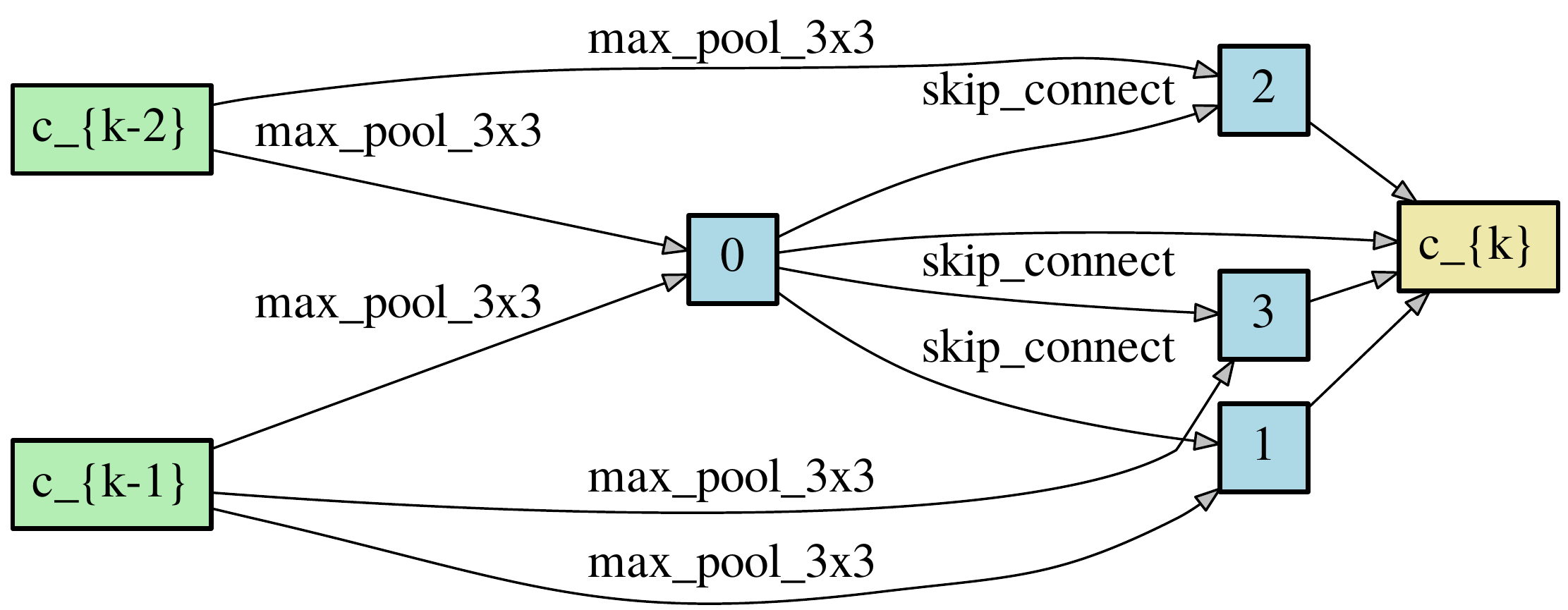}%
	}
	\subfloat[][Low-level normal cell in RC-DARTS\label{fig:rcdarts_normal_0}]{%
		\includegraphics[width=0.33\linewidth]{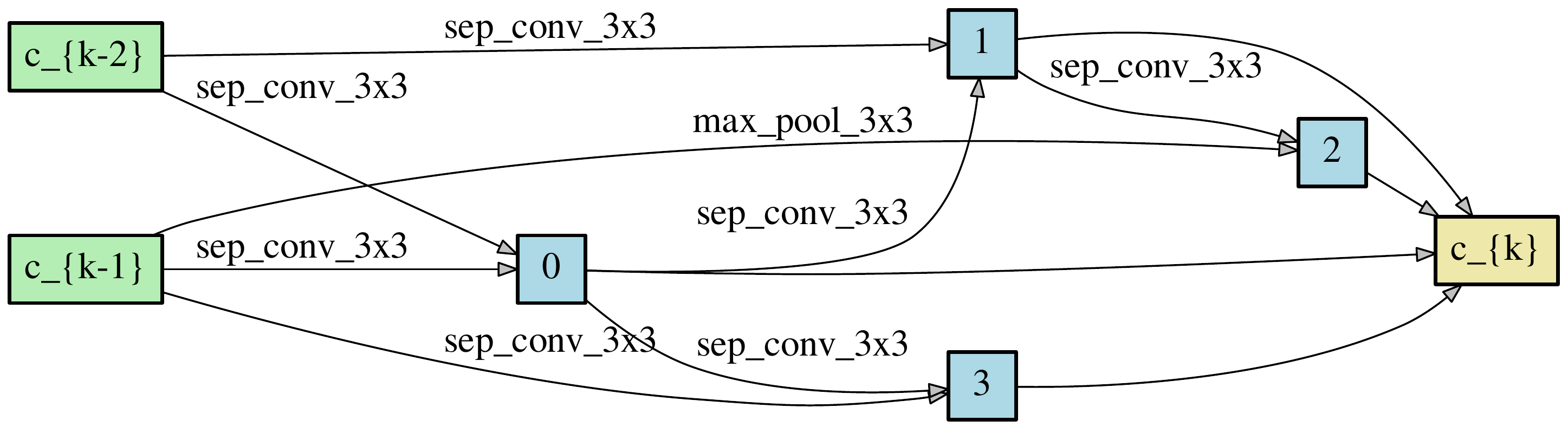}%
	} \\
	\subfloat[][Mid-level normal cell in RC-DARTS\label{fig:rcdarts_normal_1}]{%
		\includegraphics[width=0.33\linewidth]{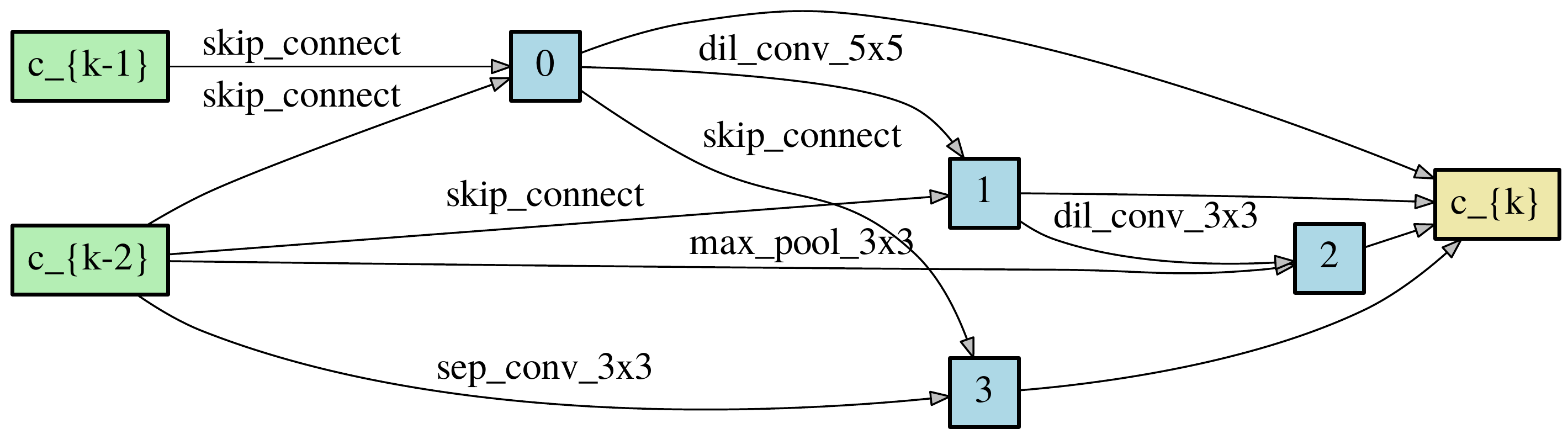}%
	}
	\subfloat[][High-level normal cell in RC-DARTS\label{fig:rcdarts_normal_2}]{%
		\includegraphics[width=0.33\linewidth]{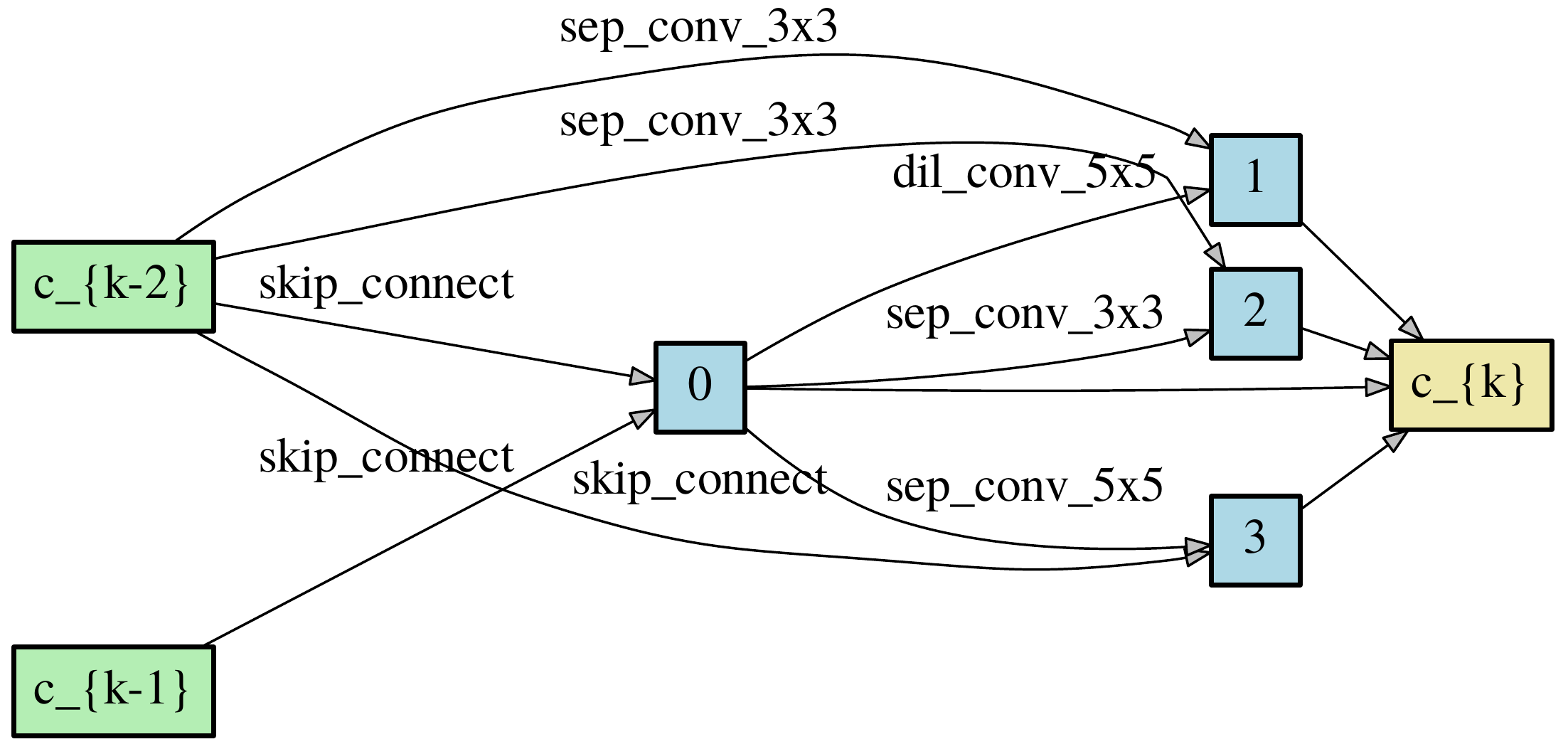}%
	}\ \ \ \ \ \ \
	\subfloat[][Reduction cell in RC-DARTS\label{fig:rcdarts_reduce}]{%
		\includegraphics[width=0.26\linewidth]{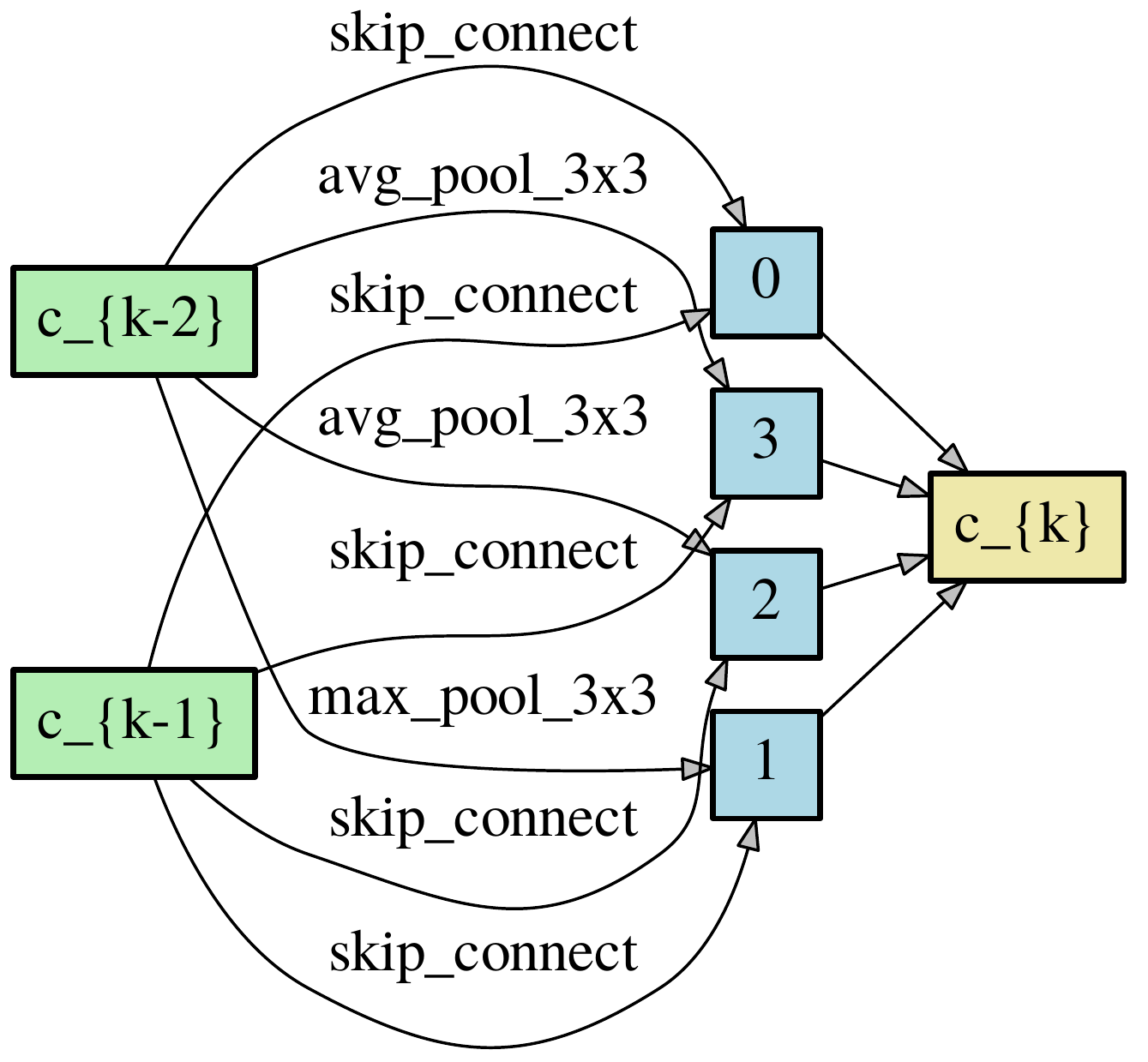}%
	}
	\caption{Comparison of cells learned in DARTS and RC-DARTS. Cells learned in DARTS are from the original paper~\cite{liu2018darts}. $c\{k-1\}$ and $c\{k-2\}$ denote the transformed outputs (by connection cell) of previous two cells before $k$th cell. RC-DARTS learns three kinds of normal cells located at low-level/mid-level/high-level of the network, respectively.
		Besides, the connection cell (omitted in the figure) learned by RC-DARTS contains one 1x1 convolution with \#gropus equals to 2.
		In comparison, DARTS predefine the connections between cells to be 1x1 normal convolution. Best viewed in zoomed in pdf.}
	\label{fig:cells_vis}
\end{figure*}

\subsection{Architecture search on Cifar10}
\subsubsection{Experimental settings} \label{sec:search-setting}
\noindent \textbf{Dataset}
Cifar10~\cite{krizhevsky2009learning} is a popular image classification benchmark containing 50,000 training images and 10,000 testing images.
During architecture search process,
we follow DARTS~\cite{liu2018darts} to evenly divide the training set into two parts,
which are used as training and validation datasets for architecture search.

\noindent\paragraph{Search space.}\quad
As described in Section~\ref{sec:mls}, there are three types of cells  in RC-DARTS:
\textit{normal}, \textit{reduction} and \textit{connection}.
Following~\cite{zoph2017learning,liu2017hierarchical,liu2017progressive,pham2018efficient,liu2018darts},
reduction cells are placed at the $1/3$ and $2/3$ of the total depth of the network between two normal cells.
Connection cells are placed between every connection of two cells to transform the
number of channels of the previous cell's output to that of the input of the current cell.

For each reduction cell and normal cell, there are $N = 7$ nodes,
including two input nodes and one output node. (See the method section for the definition of input and output nodes).
We include the following 8 types of operations in $\mathcal{O}$ for each edge at normal or reduction cell:
separable convolutions with kernel size of $3\times3$ and $5\times5$;
dilated separable convolutions with kernel size of $3\times3$ and $5\times5$;
$3\times3$ max pooling;
$3\times3$ average pooling;
identity connect (i.e. the output is equal to input);
and zero (i.e. no connection).
Both separable and dilated convolution operators are consists of ReLU-Conv-BN.
For separable convolution, two independently learned ``ReLU-Conv-BN'' on $x$ and $y$ axis are stacked.

There are four candidate operations at each edge in a  connection cells:
dilated convolution with kernel size of $3\times 3$;
$1 \times 1$ group convolutions with \#groups equivalent to 1, 2 and 4 respectively.
For group convolution with \#groups larger than one, channel shuffling~\cite{zhang2017shufflenet} is used.

\noindent\paragraph{Training settings}\quad
Following DARTS~\cite{liu2018darts}, half of the CIFAR-10 training data are held out as the validation
set for architecture search.
A small network consisting of $8$ cells is trained using RC-DARTS for 50 epochs, with batch size 64
(for both the training and validation iterations) and the initial number of channels 16.
Momentum SGD and Adam optimizer are used to optimize the weights $w$ and $\theta$ iteratively in unconstrained
training step.
In architectural projection step (ref. Eq.~\eqref{eq:project_lag}),
we use Adam optimizer with initial learning rate of 3e-4, momentum of $(0.5, 0.999)$, and no weight decay.
The number of iterations in phase I $e_u$ is set to 150 (i.e.\ 0.5 epoch);
The number of iteration in phase II $e_p$ is set to 500. The lower/upper bounds for constraints of \#params and FLOPs are 1.8e5/2.0e5 and 2.8e7/3.3e7 respectively. These numbers are decided by experiments.
All experiments run with PyTorch using NVIDIA V100 GPUs.

\subsection{Interpreting architecture search}
In Figure~\ref{fig:cells_vis}, we compare the normal and reduction cells learned by RC-DARTS with those learned by DARTS.
We have following observations as seen in the figure:
\begin{itemize}
	\item Compared with the normal cell in DARTS, the normal cells in RC-DARTS contains more inexpensive operations.
	For example, as shown in Figure~\ref{fig:rcdarts_normal_1} and Figure~\ref{fig:rcdarts_normal_2},
	There are more skip connection, max pooling in the mid-level/high-level cells in RC-DARTS.
	Compared to DARTS, RC-DARTS also uses more dilated convolution instead of separable convolution,
	because the latter is more computationally expensive.
	The above results demonstrate that RC-DARTS effectively learns lightweight architectures when resource constraints exist.
	\item Compared to DARTS, the normal cells in RC-DARTS have more connections between intermediate nodes.
	For example, in Figure~\ref{fig:darts_normal}, only the node indexed by 3 has connection with other intermediate nodes (node indexed by 2).
	In contrast, the connections among nodes in RC-DARTS is much denser, as shown in Figure~\ref{fig:rcdarts_normal_0} to Figure~\ref{fig:rcdarts_normal_1}.
	Since connections between nodes implicitly increase the depth of network,
	we conjecture that the networks with more connections between intermediate nodes have stronger learning capabilities
	when the number of parameters is the same.
	This is consistent with the expert knowledge on designing effective neural network architectures.
	For example, in the efficient inception models~\cite{szegedy2016rethinking,szegedy2017inception}, there are rich connections inside each inception module.
\end{itemize}

\subsection{Architecture evaluation}
\subsubsection{Cifar10}
\paragraph{Experimental settings}
For a fair comparison with DARTS, we follow~\cite{liu2018darts} to build large network of $20$ cells, which is trained for 600 epochs with batch size 96.
Other training hyperparameters is also the same  DARTS.
Following ~\cite{pham2018efficient,zoph2017learning,liu2017hierarchical,real2018regularized},
we add additional improvements include cutout~\cite{devries2017improved}, path dropout of probability 0.3 and auxiliary towers with weight 0.4.
We report the mean results of 4 independent runs of training our full model.
In order to compare with DARTS with the same number of parameters, we adjust the multiplier of \#channels for both DARTS and RC-DARTS to make the same model size of them roughly the same.
The resulted models are denoted by ``\{model-C\#channels\}''.

\paragraph{Results} The comparison of RC-DARTS with other state-of-the-arts on CIFAR-10 are presented in Table~\ref{table:cifar10}. Notably, RC-DARTS
achieved comparable results with the state of the art~\cite{zoph2017learning, real2018regularized} while using
three orders of magnitude less computation resources (i.e. 1 GPU day vs 1800 GPU days for
NASNet and 3150 GPU days for AmoebaNet). Compared with full-sized DARTS in~\cite{liu2018darts} which has an initial number of channels 16, RC-DARTS-C42 achieves competitive performance but with less parameters and less search cost (1 versus 4). Since RC-DARTS aims to learn lightweight models under resource constraints, we put more emphasis on the comparison results of models with low resource occupations. Compared with compressed DARTS, (i.e. DARTS-C20 and DARTS-C12), RC-DARTS-C22 and RC-DARTS-C14 outperforms them by a large margin, which evidently demonstrates the effectiveness of RC-DARTS on learning resource constrained architectures.

Compared with DPPNet, as shown in Table 2, on Cifar10, RC-DARTS-C14 largely reduces the error rate of DPPNet by 1.67\% (a 28.6\% relative improvements). 
We use RC-DARTS-C54 model for clearer comparions with DPPNet on ImageNet. 
With similar FLOPs (520M vs 523M), RC-DARTS-C54 achieves lower error rate (25.7\% vs 26.0\%), smaller model size (4.4M vs 4.8M) and search costs (1 vs 8 GPU days). 
Above results demonstrate the advantages of RC-DARTS over DPPNet.
\begin{table*}[t]
	\footnotesize
	\cprotect\caption{Comparison with state-of-the-art image classifiers on CIFAR-10. Models with symbol of $\dagger$ are run by us using authors' realeased implementations. The number in \{model-C\#\} is the initial number of channels. Both DARTS and RC-DARTS models use cutout in training. ER stands for ``error rate''. }
	\centering
	\begin{tabular}{lcccc} \toprule Architectures    &Test Error (\%)   & Params (M)  & Search Cost (GPU days) & Search Method   \\
		\midrule
		DenseNet-BC~\cite{huang2017densely}                 & 3.46 & 25.6 & - & mannual \\
		\midrule
		NASNet-A+cutout~\cite{zoph2017learning}    & 2.65   & 3.3 & 1800 & RL      \\
		NASNet-A  +cutout~\cite{zoph2017learning} & 2.83 & 3.1 & 3150 & RL \\
		AmoebaNet-A + cutout~\cite{real2018regularized} & 3.34 $\pm$ 0.06 & 3.2 & 3150 & evolution \\
		AmoebaNet-A + cutout~\cite{real2018regularized} & 3.12 & 3.1 & 3150 & evolution \\
		AmoebaNet-B + cutout~\cite{real2018regularized} & 2.55 $\pm$ \ 0.05 & 2.8 & 3150 & evolution \\
		Hierarchical Evo~\cite{liu2017hierarchical} & 3.75 $\pm$ 0.12 & 15.7 & 300 & evolution \\
		PNAS~\cite{liu2017progressive}            &             3.41 $\pm$ 0.09 & 3.2 & 225 & SMBO \\
		ENAS + cutout~\cite{pham2018efficient} & 2.89 & 4.6 & 0.5 & RL \\
		DPPNet~\cite{dong2018dpp}        &      5.84  & 0.45 & 8 & RL \\
		\midrule
		DARTS~\cite{liu2018darts}        &        2.83 $\pm$ 0.06 & 3.4 & 4 & gradient-based \\
		DARTS$^\dagger$-C12        &      3.44 &        1.0          &          4       & gradient-based \\
		DARTS$^\dagger$-C20        &         4.86          &           0.48        &      4 & gradient-based \\
		RC-DARTS-C42        &        2.81 $\pm$ 0.03  &    3.3        &      1     & gradient-based \\
		RC-DARTS-C22        &        3.02  &    1.0        &      1     & gradient-based   \\
		RC-DARTS-C14       &        4.17   &  0.43    &  1     & gradient-based   \\
		\bottomrule
		\label{table:cifar10}
	\end{tabular}
\end{table*}

\begin{table*}[t]
	\footnotesize
	\cprotect\caption{Comparison with state-of-the-art image classifiers on ImageNet. The Models with symbol of $\dagger$ are run by us using authors' realeased implementations. The number in \{model-C\#\} is the initial number of channels. ER stands for ``error rate''. }
	\centering
	\begin{tabular}{lcccccc} \toprule
		\multirow{2}{*}{Architectures}   & \multicolumn{2}{c@{}}{Error rates (\%)}  &  \multirow{2}{*}{Params (M)}& \multirow{2}{*}{FLOPs (M)}& \multirow{2}{*}{Search Cost} & \multirow{2}{*}{Search Method}\\
		\cmidrule(lr){2-3}
		&Top1 & Top5   & &  & (GPU days) &    \\
		\midrule
		Incepton-V1~\cite{szegedy2015going} & 30.2 & 10.1 & 6.6 & 1448 & - & manual \\
		MobileNet~\cite{howard2017mobilenets} & 29.4 & 10.5 & 4.2 & 569 & - & manual \\
		ShuffleNet 2$\times$ (v1)~\cite{zhang2017shufflenet} & 29.1 & 10.2 & $\sim$5 & 524 & - & manual \\
		ShuffleNet 2$\times$ (v1)~\cite{zhang2017shufflenet} & 26.3 & - & $\sim$5 & 524 & -  & manual \\
		\midrule
		MnasNet~\cite{tan2018mnasnet} & 26.0 & 8.25 & 4.2 & 317 & 1800 & RL \\

		NASNet-A~\cite{zoph2017learning}    & 26.0 & 8.4 & 5.3 & 564 & 1800  & RL \\
		NASNet-B~\cite{zoph2017learning}    & 27.2 & 8.7 & 5.3 & 488 & 1800 & RL \\
		NASNet-C~\cite{zoph2017learning}    & 27.5 & 9.0 & 4.9 & 558 & 1800 & RL \\
		AmoebaNet-A~\cite{real2018regularized} & 25.5 & 8.0 & 5.1 & 555 & 3150 & evolution \\
		AmoebaNet-B~\cite{real2018regularized} & 26.0 & 8.5 & 5.3 & 555 & 3150 & evolution \\
		AmoebaNet-C~\cite{real2018regularized} & 24.3 & 7.6 & 6.4 & 570 & 3150 & evolution \\
		PNAS~\cite{liu2017progressive}                &  25.8 & 8.1 & 5.1 & 588 & $\sim$225 & SMBO \\
		DPPNet~\cite{dong2018dpp}        &   26.0 & 8.2 & 4.8 & 523 & 8 & SMBO  \\
		\midrule
		DARTS            &   26.9 & 9.0 & 4.9 & 595 & 4 & gradient-based \\
		DARTS$^\dagger$-C24        & 34.8 &  13.8 & 1.4 & 140 & 4 & gradient-based \\
		RC-DARTS-C58        &  25.1  &    7.8  &    4.9        &  590 &    1     & gradient-based   \\
		RC-DARTS-C28        &  32.9  &   13.3  &    1.4        &  138 &    1     & gradient-based   \\
		\bottomrule
		\label{table:imagenet}
	\end{tabular}
\end{table*}

\subsubsection{ImageNet}
\paragraph{Experimental settings}  A network of 14 cells is trained for 250
epochs with batch size 128, weight decay 3e-5
and initial SGD learning rate 0.1 (decayed by a
factor of 0.97 after each epoch). Other hyperparameters follow~\cite{zoph2017learning,real2018regularized,liu2018darts}. The training takes 4 days on a single GPU.

\paragraph{Ablative study} 	
We evaluate the effects of MLS and CC on DARTS and RC-DARTS based on four combinations (with/without MLS/CC).
We adjust the multiplier of \#channels of different models to compare them with comparative model sizes. 
Two model sizes (4.9M and 1.4M) are used for evaluating models with different resource costs. All other training settings are the same as in Sec.~\ref{sec:search-setting}. The average values of four independent runs are reported to reduce randomness.

\begin{table}[t]
	\small
	\cprotect\caption{Comparison results of different settings of the proposed multi-level search (MLS) and connection cell (CC) on DARTS and RC-DARTS. For brevity, the \#channels corresponding to each model is omited.}
	\centering
	\begin{tabular}{l@{\hskip 0.3cm}c@{\hskip 0.3cm}c@{\hskip 0.3cm}c@{\hskip 0.3cm}c@{\hskip 0.3cm}c} \toprule \specialrule{0.0em}{0.pt}{2pt}
		\multirow{2}[2]{*}{Architectures}   & \multicolumn{4}{c}{Error rates (\%) on ImageNet test set}  &  \multirow{2}{*}{Params} \\
		\specialrule{0.0em}{0.5pt}{0pt}\cmidrule(r){2-5}\specialrule{0.0em}{0pt}{-0.2pt}
		&w/o (MLS+CC) & w/ MLS   &  w/ CC   & w/ (MLS+CC)  & (M) \\ 
		\specialrule{0.0em}{1pt}{0pt}\midrule \specialrule{0.0em}{0pt}{2pt}
		DARTS        &  26.9 &  29.3  &  26.4  & 28.6 & 4.9 \\
		RC-DARTS  &  26.0  &  25.6  &  25.4  & \textbf{25.1} & 4.9 \\
		\specialrule{0.05em}{0.5pt}{2pt}
		DARTS        &  34.8  &  37.7 &  34.3  & 36.2 & 1.4 \\
		RC-DARTS  &  33.7  &  33.4  &  33.5  & \textbf{32.9 }& 1.4 \\
		\specialrule{0.0em}{0.6pt}{0pt}\bottomrule
		\label{table:ablation}
		\vspace{-0.8mm}
	\end{tabular}
	\par
\end{table}

When none of MLS and CC is used, compared with DARTS with similar model sizes (4.9M/1.4M), RC-DARTS reduce the error rates by 0.9\%/1.1\% respectively.
%
These results verify the effectiveness of the proposed iterative projection algorithm. 
By using MLS, the performance of RC-DARTS is improved as it adaptively learns architectures at different layers (see Sec.~\ref{sec:mls}). 
Interestingly, we find MLS incurs accuracy loss for DARTS. 
This may be due to the overfitting problem caused by the increased architecture parameter $\theta$ (note its size is proportional to \#levels in MLS). 
Experimental results verify this point as compared with RC-DARTS, DARTS has higher error rates in test set but lower error rates in training set. 
In contrast, the resource constrains (i.e. Eq.~\eqref{eq:lnas_3}) in RC-DARTS can be seen as regularizers for the parameter space, thereby reducing the risk of overfitting. 
%
Using CC consistently reduces error rates of both DARTS and RC-DARTS.  
Finally, RC-DARTS with both MLS and CC achieves the lowest error rates among all models. 

\paragraph{Results }
Table~\ref{table:imagenet} lists the comparison results of the evaluation on ImageNet dataset.
Results in Table 2 show that the cell learned on CIFAR-10 is transferable to ImageNet.
Notably, RC-DARTS achieves competitive performance with the state-of-the-art RL method~\cite{zoph2017learning} while using three orders of magnitude less computation resources for architecture training.
Compared with DARTS with the same number of parameters and FLOPs, RC-DARTS significantly outperforms DARTS in terms of accuracy.

Compared with MobileNetV2x1.4, RC-DARTS-C58 achieves the same error rate and comparable FLOPs, but using only 71\% of the former's model size (4.9M vs 6.9M). Besides, we also compare RC-DARTS-C46 (i.e. set the \#channels to 46) with MobileNetV2. The former achieves: top1 error rate=27.6\%, FLOPs=380M and \#params=3.2.   
It thus can be seen that with comparable FLOPs, RC-DARTS-C46 achieves lower error rate with smaller model size. Above experiments demonstrate the advantage of RC-DARTS over MobileNetV2 in learning lightweight models. Although MNasNet-92 has smaller FLOPs vs RC-DARTS-C58 with comparable error rate, the former takes \textgreater2000 times more search costs than the latter, thus prohibiting it from being used in computational budgets limited circumstances.

Lastly, we note both MobileNetV2 and MNasNet use the more efficient inverted residual block (IRB)~\cite{mnetv2} as basic block/cell. Since the set of candidate operations in RC-DARTS is orthogonal to our proposed methods, we can conveniently replace the ordinary convolutional operations in RC-DARTS with IRB to boost the performance. Since it is out of the scope of this paper (i.e. learning one-shot NAS with resource constraints), we plan to leave this investigation to our future work.
Although RC-DARTS has larger FLOPs than MNasnet~\cite{tan2018mnasnet} with comparable error rate, but the search cost of former is an order of magnitude smaller~\cite{tan2018mnasnet}, which is essential for one-shot architecture search.

\section{Related Work}
Recently, automatic deep neural network architecture search (NAS) has been demonstrated effective in multiple AI tasks by achieving the state-of-the-art results. We classify those works on NAS in three different types according to the employed optimization approaches: reinforcement learning based methods, evolutionary algorithm based ones and methods using other optimization approaches. In addition, we also review works that contain multiple objectives in optimization and manually designed mobile architectures.
\paragraph{RL-based methods}In the seminal work of \cite{zoph2016neural}, a Neural
Architecture Search (NAS) method is developed to search network architectures. NAS consists of two basic component, \textit{i.e.} a LSTM-based controller which aims to generate layer-wise architecture options (\textit{e.g.} filter size/stride, pooling etc.) and the REINFORCE algorithm~\cite{williams1992simple} which updates the weights of controller using the accuracy of sampled architectures as rewards. Many works follow using the similar pipeline. \cite{zoph2017learning} proposes NASNet by introducing a modular search space to reduce the search cost and enhance the generalization capability of searched architectures. Specifically, the basic architectural building block (named ``cell'' therein) is first learnt in small-scale dataset and then transfered to large-scale dataset by stacking together multiple copies. ENAS~\cite{pham2018efficient} speeds up NAS by sharing parameters across child models during the architectural search process. It is observed that such a weight sharing scheme also improve the performance of searched models. \cite{cai2018efficient} proposes an efficient architecture search method by restricting the actions output by controller to be varying the depth/width of network with function-preserving transformations. In such a way, the previously validated child model can be reused to save search cost. Instead of using REINFORCE, \cite{baker2016designing} and
~\cite{zhong2017practical} use Q-learning in architecture search. Above methods focus on improving the accuracy of searched models and ignore the resource cost of models. In comparison, RC-DARTS aims to maximize accuracy under resource constraints, thus being more suitable for learning lightweight architectures in mobile platforms.

More recently, there are several works integrating resource constraints as objectives in architecture searching.~\cite{RENA}	modifies the reward function to penalize searched models when constraints are not met. The method is only tested on small-scale dataset. MNasNet~\cite{tan2018mnasnet} explictly adds resource constraints in rewards and achieves promising results in trading-off accuracy and model complexity. Compared with MNasNet, RC-DARTS achieves comparable performance under similiar resource constraints, but using only three orders less search time (see experimental section). 
\paragraph{EA-based methods} Early attemps using evolutionary algorithm to optimize neural architecture includes~\cite{angeline1994evolutionary,stanley2002evolving,floreano2008neuroevolution}. However, since EA is used for learning both architectures and weights, these methods are difficult to be applied in modern deep networks which have a large amount of parameters. Recent works including~\cite{real2017large,xie2017genetic,liu2017hierarchical,real2018regularized} separate the learning of architecture and weights, \textit{i.e.} the former is learned using EA while the later is learned using conventional gradient descent. During the process of learning architecture using EA, ``mutation'' represents operations to choose different architectural options, \textit{e.g.} the filter size, layer number, etc. [19] proposed to treat neural network architecture search as a multi-objective optimization task and adopt an
evolutionary algorithm to search models with two objectives, run-time speed,
and classification accuracy. However, the performances of the searched models
are not comparable to handcrafted small CNNs, and the numbers of GPUs they
required are enormous. A prominent disadvantage of EA-based methods is that they require enormous computational resources (generally a few hundreds of GPU days), thus prohibiting its usage in budget limited circumstances.

\paragraph{Other optimization-based methods}  A gaussian process-based bayesian optimization is used in~\cite{kandasamy2018neural,swersky2014raiders} for neural architecture search. However, its performance is unsatisfying compared with previous two categories of methods. To reduce the excessive computation costs on evaluating candidate architectures in original NAS method~\cite{zoph2016neural}, \cite{baker2017accelerating} proposes
to predict performance of model architectures instead of performing conventional training. ~\cite{brock2017smash} adopts a similar idea by  training an extra network to generate the weights of
candidate architectures and using random search to search for good
models. In another direction, the sequential model-based optimization, i.e. SMBO~\cite{hutter2011sequential} algorithm is adopted to guide the selection of model architectures through learning a predictive model. Based on SMBO, \cite{liu2017progressive} achieves comparable performance to NASNet with significantly smaller computational budget. Following~\cite{liu2017progressive},  DPPNet~\cite{dong2018dpp} integrates resource constraints in the objective function for searching efficient models. Compared with DPP-Net, since RC-DARTS performs network architecture search in a one-shot manner, thus RC-DARTS not only has higher searching efficiency, but also achieves better trade-off between accuracy and model complexity as validated by extensive experimental results. 

Recently, \cite{liu2018darts} solves the NAS problem from a different angle, and proposes a method for efficient
architecture search named DARTS (Differentiable Architecture Search). Instead of searching over
a discrete set of candidate architectures, they relax the search space to be continuous, so that the
architecture can be optimized with respect to its validation set performance by gradient descent. However, above methods ignores the computational costs of obtained architectures, making the learned models sub-optimal in computational resource usages. RC-DARTS is built upon DARTS, but with significant improvement in the effeciency of learned models with better accuracies. Different from DARTS, 
RC-DARTS aims to learn optimal architectures that satisfy customized resource constrains. 
To the best of our knowledge, RC-DARTS is the first to resolve one-shot NAS with resource constraints. \textbf{(2)} As the exact resource costs of output architectures are discrete, it is challenging to incorporate resource constraints into training and solve the optimization problem. 
We approximate the resource costs with continuous relaxation and address the non-convex constrained optimization with a novel iterative projection algorithm. 
In addition, we propose multi-level search strategy and new types of connection cell to achieve better trade-off between accuracy and resource costs. 

\paragraph{Handcrafted lightweight models} There have been a lot of handcrafted models proposed to achieve high performance on mobile devices with limited computational resources. Among these lightweight models, group convolution and its variants play critical roles in improving models' efficiency. The IGCV models~\cite{IGCV1,IGCV2} introduce novel interleaved group convolutions to improve the representation capablilities of models with the same number of parameters and model complexities. Both MobileNet~\cite{howard2017mobilenets} and
ShuffleNet~\cite{zhang2017shufflenet} employ depth-wise convolution (an extreme case of group convolution when the number of groups is equal to the input's number of channels) to significantly reduce model sizes and complexities while retaining comparable accurate. CondenseNet~\cite{condensenet} proposes a novel learnable group convolution to improve the computational efficiency of DenseNet. MobileNet V2~\cite{mnetv2} proposes an inverted residual
block (IRB) to achieve better trade-off between accuracy and computational costs than~\cite{howard2017mobilenets}. 
ShuffleNet V2~\cite{shufflenetv2} proposes pratical guidelines for designing efficient networks and achieves state-of-the-art results in terms of efficiency and accuracy trade-off. Although above models achieve promising results in balancing models' accuracy and computational costs, their design process are usually time-consuming and costly and are subject to trial-and-errors by experts in the field. In comparison, our proposed RC-DARTS is a general architecture search method which can automatically learn architectures from data in a one-shot way. More importantly, we experimentally verify that RC-DARTS is advantageous over handcrafted models in the synthetical criteria of model size/complexity, search costs and accuracy. 

\section{Conclusion}
In this work, we presented RC-DARTS, a novel end-to-end neural architecture search
framework, for one-shot neural architecture search under resource constraints, where
a customized network architecture is learned for any particular dataset.
RC-DARTS employs differentiable architectural techniques while taking the resource constraints
into consideration by adding them as constraints in the objective function.
RC-DARTS achieves a good balance between architecture search speed, resource constraints, and
quality of the architecture. It is ideal for the resource constrained neural architecture search problem.
On Cifar10 and ImageNet datasets,
RC-DARTS achieves state-of-the-arts performance in terms of accuracy, model size and complexity.
In the future, we plan to explore using neural network to learn the mapping function between
network hyperparameters and a more complicated resource cost, such as inference speed on specific hardware.

\bibliographystyle{plain}
\bibliography{mybib}

\end{document}